\pdfoutput=1
\documentclass{llncs}
\usepackage{llncsdoc}

\usepackage{amsmath}
\usepackage{amsfonts}
\usepackage{amscd}
\usepackage{amssymb}
\usepackage{mathrsfs}
\usepackage{bbm}
\usepackage[dvipdfmx]{graphicx}
\usepackage{color}
\usepackage{here}

\def\RR{\mathbb{R}}
\usepackage{color}
\def\RR{\mathbb{R}}
\definecolor{R}{rgb}{1., 0., 0.}

\begin{document}
%

\title{
Principal Sensitivity Analysis
}

\author{
Sotetsu Koyamada\inst{1}\inst{2}
Masanori Koyama\inst{1}
Ken Nakae\inst{1}
\and Shin Ishii\inst{1}\inst{2}}

\institute{
Graduate School of Informatics, Kyoto University, Kyoto, Japan
\and
ATR Cognitive Mechanisms Laboratories, Kyoto, Japan
\email{koyamada-s@sys.i.kyoto-u.ac.jp, ishii@i.kyoto-u.ac.jp}
}



\maketitle
%
%
%
%
%
\begin{abstract}
 We present a novel algorithm (Principal Sensitivity Analysis; PSA)
 to analyze the knowledge of the classifier
 obtained from supervised machine learning techniques.
 In particular, we define principal sensitivity map (PSM) as the
 direction on the input space to which the trained classifier is most
 sensitive, and use analogously defined $k$-th PSM to define a basis for
 the input space.  We train neural networks with artificial data and
 real data, and apply the algorithm to the obtained supervised classifiers.
 We then visualize the PSMs to demonstrate the PSA's ability to decompose
 the knowledge acquired by the trained classifiers.
\end{abstract}

\section{Introduction}
Machine learning is a powerful methodology to construct efficient and robust
predictors and classifiers.
Literature suggests its ability in the supervised context not only to reproduce
``intuition and experience'' based on human supervision~\cite{DeepFace},
but also to successfully classify the objects that humans cannot
sufficiently classify with inspection alone~\cite{Horikawa2013,Mural1991}.
This work is motivated by the cases in which the machine
classifier eclipses the human decisions.
We may say that this is the case in which the
classifier holds more knowledge about the classes than us, because our
incompetence in the classification problems can be attributed solely to
our lack of understanding about the class properties and/or the similarity metrics.
The superiority of nonlinear machine learning techniques strongly suggests that the trained classifiers
capture the ``invisible'' properties of the subject classes.
Geoff Hinton solidified this into the philosophy of
``dark knowledge'' captured within the trained classifiers~\cite{DarkKnowledge}.
One might therefore be motivated to enhance understanding of subject
classes by studying the way the trained machine acquires the
information.

Unfortunately, trained classifiers are often so complex that they defy human interpretation.
Although some efforts have been made to ``visualize'' the
classifiers~\cite{Baehrens2010,Rasmussen2011},
there is still much room left for improvement.
The machine learning techniques in neuroimaging, for example,
prefer linear kernels to nonlinear kernels because of the lack of
visualization techniques~\cite{LaConte2005}.
For the visualization of high-dimensional feature space of machine learners,
Zurada et al.~\cite{Zurada1994,Zurada1997} and Kjems et al.~\cite{Kjems2002}
presented seminal works.
Zurada et al.\ developed ``sensitivity analysis'' in order to ``delete
unimportant data components for feedforward neural networks.''
Kjems et al.\ visualized Zurada's idea as ``sensitivity map''
in the context of neuroimaging.
In this study, we attempt to  generalize the idea of sensitivity
analysis, and develop a new framework that aids us in extracting the
knowledge from classifiers that are trained in a supervised manner.
Our framework is superior to the predecessors in that it can:
\begin{enumerate}
 \item be used to identify a pair of discriminative input
       features that act oppositely in characterizing a class,
 \item identify \textit{combinations} of discriminative features that strongly
       characterize the subject classes,
 \item provide platform for developing sparse, visually intuitive
       sensitivity maps.
\end{enumerate}
The new framework gives rise to the algorithm that we refer
to as ``Principal Sensitivity Analysis (PSA),'' which is analogous to
the well-established Principal Component Analysis (PCA).

\section{Methods}
%
\subsection{Conventional sensitivity analysis}
%
Before introducing the PSA, we describe the original sensitivity map
introduced in~\cite{Kjems2002}.
Let $d$ be the dimension of the input space, and let
$f: \RR^d \to \RR$ be the classifier function obtained from supervised training.
In the case of SVM, $f$ may be the discriminant function.
In the case of nonlinear neural networks, $f$ may represent the function (or log of
the function) that maps the input to the output of a unit in the final layer.
We are interested in the expected sensitivity of $f$
with respect to the $i$-th input feature.
This can be written as
\begin{align}
\begin{split}
s_i := \int \left( \frac{\partial f(\vec{x})}{\partial
 x_i} \right)^2 q(\vec{x}) d\vec{x}
\end{split}  \label{eq:original_sensitivity},
\end{align}
where $q$ is the distribution over the input space.
In actual implementation, the integral~\eqref{eq:original_sensitivity} is
computed with the empirical distribution $q$ of the test dataset.
Now, the vector
\begin{align}
\begin{split}
\vec{s} := \left(s_1, \dots, s_d \right)
\end{split} \label{eq:classical_def}
\end{align}
of these values will give us an intuitive measure for the degree
of importance that the classifier attaches to each input.
Kjems et al.~\cite{Kjems2002} defined $\vec{s}$ as \textbf{sensitivity map} over the set
of input features.
%
\subsection{Sensitivity in arbitrary direction}
%
Here, we generalize the definition~\eqref{eq:original_sensitivity}.
We define $s(\vec{v})$ as the sensitivity of $f$ in arbitrary direction $\vec{v} := \sum_{i}^{d} v_i
\vec{e}_i$, where $\vec{e}_i$ denotes the $i$-th standard basis in $\RR^d$:
%
\begin{align}
\begin{split}
 s(\vec{v}) := \int \left( \frac{\partial f(\vec{x})}{\partial \vec{v}} \right)^2 q(\vec{x}) \, d\vec{x}.
\end{split} \label{eq:koyamada_sensitivity}
\end{align}
Recall that the directional derivative is defined by
\[
 \frac{\partial f(\vec{x})}{\partial \vec{v}} := \sum_{i=1}^{d} v_i \frac{\partial f(\vec{x})}{\partial x_i}.
\]
Note that when we define the \textit{sensitivity inner product}
\begin{align}
\begin{split}
 \langle \vec{e}_i, \vec{e}_j \rangle_{s} := \int
\left( \frac{\partial f(\vec{x})}{\partial x_i} \right)
\left( \frac{\partial f(\vec{x})}{\partial x_j} \right)
q(\vec{x}) \, d\vec{x},
\end{split}  \label{eq:koyamada_inner_p}
\end{align}
we can rewrite $s(\vec{v})$ with the corresponding \textit{sensitivity norm}, as follows:
\begin{align}
\begin{split}
\|\vec{v}\|^2_{s} &:= \langle \vec{v}, \vec{v} \rangle_{s} \\
 &= \left\langle \sum_{i} v_i \vec{e}_i ,\, \sum_{j} v_j \vec{e}_j \right\rangle_{s} \\
 &= \sum_{i, j} v_i v_j \left \langle \vec{e}_i , \vec{e}_j
 \right\rangle_{s}.
\end{split}  \label{eq:koyamada_inner_p2}
\end{align}
This inner product defines the kernel metric
corresponding to the positive definite matrix $\vec{K}$ with $ij$-th entry
given by  $K_{ij} :=  \left\langle \vec{e}_i , \vec{e}_j \right\rangle_{s}$.
This allows us to write
\begin{align}
\begin{split}
 s(\vec{v}) = \vec{v}^{\mathrm{T}} \vec{K} \vec{v}.
\end{split} \label{eq:koyamada_metric}
\end{align}
\subsection{Principal sensitivity map and PSA}
The classical setting~\eqref{eq:classical_def} was developed
in order to quantify the sensitivity of $f$ with respect
to each individual input feature.
We attempt to generalize this idea and seek the \textit{combination
of the input features} for which $f$ is most sensitive, or the combination of the input features that is
\textit{``principal''} in the evaluation of the sensitivity of $f$.
We can quantify such combination by the vector $\vec{v}$, solving the following
optimization problem about $\vec{v}$:
\begin{align}
  \begin{split}
\begin{aligned}
& \text{maximize}
& & \vec{v}^{\mathrm{T}} \vec{K} \vec{v} \\
& \text{subject to}
& & \vec{v}^{\mathrm{T}}\vec{v} = 1.
\end{aligned}
 \end{split} \label{def:psm}
\end{align}
The solution to this problem is simply the maximal eigenvector
$\pm\vec{v}^{*}$ of $\vec{K}$.
Note that $v_i$ represents the contribution of the $i$-th input feature to this principal
combination, and this gives rise to the map over the set of all input features.
As such, we can say that $\vec{v}$ is the \textbf{principal sensitivity map
(PSM)} over the set of input features.
From now on, we call $\vec{s}$ in the classical
definition~\eqref{eq:classical_def} as the \textbf{standard sensitivity
map} and make the distinction.
%
The magnitude of $v_i$ represents the extent
to which $f$ is sensitive to the $i$-th input feature,
and the sign of $v_i$ will tell us the relative direction
to which the input feature influences $f$.
The new map is thus richer in information than the standard
sensitivity map.
In Section 3.1 we will demonstrate the benefit of this extra information.
\subsubsection{Principal Sensitivity Analysis (PSA)}
We can naturally extend our construction above and also consider other
eigenvectors of $\vec{K}$.
We can find these vectors by solving the following optimization
problem about $\vec{V}$:
\begin{align}
\begin{split}
\begin{aligned}
& \text{maximize}
& & \mathrm{Tr}\left(\vec{V}^{\mathrm{T}} \vec{K} \vec{V}\right) \\
& \text{subject to}
& & \vec{v}_i^{\mathrm{T}}\vec{v}_j = \delta_{ij},
\end{aligned}
\end{split} \label{eq:PSA}
\end{align}
where $\vec{V}$ is a $d \times d$ matrix. As is well known, such $\vec{V}$ is given by
the invertible matrix with each column corresponding to $\vec{K}$'s
eigenvector.
We may define $k$-th dominant eigenvector $\vec{v}_{k}$ as
the \textbf{$k$-th principal sensitivity map.} These sub-principal sensitivity maps grant us
access to even richer information that underlies the dataset.
We will show the benefits of these additional maps in Fig.\,3.
From now on, we will refer to the first PSM by just PSM, unless noted otherwise.

Recall that, in the ordinary PCA,  $\vec{K}$ in~\eqref{eq:PSA} is given by the covariance
$E\left[ \vec{x} \vec{x}^{\mathrm{T}} \right]$, where $\vec{x}$ is the
centered random variable.
Note that in our particular case, if we put
\begin{align}
\begin{split}
\vec{r}(\vec{x}) :=
\left(
\left( \frac{\partial f(\vec{x})}{\partial x_1} \right),
\dots ,
\left( \frac{\partial f(\vec{x})}{\partial x_d} \right)
\right)^{\mathrm{T}},
\end{split}
\end{align}
then we may write
$\vec{K} = \int \vec{r(x)}\vec{r(x)}^{\mathrm{T}}
q(\vec{x}) \, d\vec{x} = E\left[ \vec{r(x)}\vec{r(x)}^{\mathrm{T}} \right]. $
We see that our algorithm can thus be seen as the PCA
applied to the covariance of $\vec{r}(\vec{x})$ without centering.
%
%
\subsubsection{Sparse PSA}
One may use the new definition~\eqref{eq:PSA} as a starting
point to develop sparse, visually intuitive sensitivity maps.
For example, we may introduce the existing techniques in sparse PCA and
sparse coding into our framework.
We may do so~\cite{Jenatton2009} by replacing the covariance matrix in
its derivation with our $\vec{K}$.
In particular, we can define an alternative optimization problem about
$\vec{V}$ and $\vec{\alpha}_i$:
\begin{align}
\begin{split}
 \begin{aligned}
& \text{minimize}
& & \frac{1}{2} \sum_{i}^{N} \| \vec{r} \left(\vec{x}_i \right)
  - \vec{V} \vec{\alpha}_i \|_2^2 + \lambda \sum_{k}^{p}
  \|\vec{v}_k\|_1\\
& \text{subject to}
& & \|\vec{\alpha}_i\|_2 = 1,
 \end{aligned}
\end{split}
\label{eq:sparsePSA}
\end{align}
where $p$ is the number of sensitivity maps and $N$ is the number of samples.
For the implementation, we used scikit-learn~\cite{Pedregosa2012}.
\subsection{Experiments}
\begin{figure}[htbp]
 \centering
 \includegraphics[width=10cm]{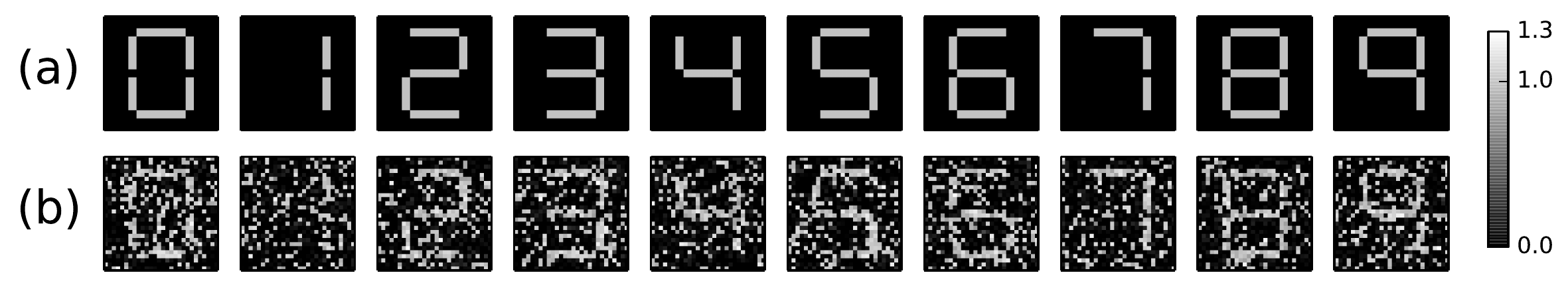}
 \caption{
 (a) Templates. (b) Noisy samples. Each figure is of $28 \times 28$ pixels.
 }
 \label{fig:1}
\end{figure}

In order to demonstrate the effectiveness of the PSA,
we applied the analysis to the classifiers that we trained with
artificial data and MNIST data.
Our artificial data is a simplified version of the MNIST data in which the object's
\textit{orientation} and \textit{positioning} are registered from the beginning.
All samples in the artificial data are constructed by adding noises to
the common set of templates representing the numerics
from $0$ through $9$ (Fig.\,\ref{fig:1}).
We then fabricated the artificial noise in three steps:
we
(1) flipped the bit of each pixel in the template picture with probability $p = 0.2$,
(2) added Gaussian noise $\mathcal{N}(0,0.1)$ to the intensity, and
(3) truncated the negative intensities.
The sample size was set to be equal to that of MNIST. Our training data,
validation data, and test data consisted respectively of 50,000, 10,000,
and 10,000 sample patterns.
Using the artificial dataset above and the standard MNIST, we trained a feed forward neural
network for the classification of ten numerics.
In Table~1, we provide the structure of the neural network and its performance over
each dataset.
For either dataset, the training was conducted via stochastic gradient
descent with constant learning rate.
%
We also adopted a dropout method~\cite{Hinton2012} only for the training
on the MNIST dataset.
The output from each unit in the final layer is given by the posterior
probability of each class $c$.
For computational purpose, we transform this output by $\log$:
\begin{align}
 f_{c}(\vec{x}) := \log P(Y = c \,|\, \vec{x}),
\end{align}
where $Y$ is, in the model governing the neural network, a random variable
representing the class that the classifier assigns to the input $\vec{x}$.
We then constructed the PSM and the standard sensitivity map for the $f_c$
given above.
\begin{table}
 \caption{Summary of training setups based on neural networks}
 \centering
 \begin{tabular}{lccccc}
  \hline
  \textbf{Data set}\,\,\, & \textbf{Architecture}\,\,\, & \textbf{Unit type}\,\,\, & \textbf{Dropout}\,\,\,
  & \textbf{Learning rate}\,\,\, & \textbf{Error[\%]} \\
  \hline
  Digital data & 784-500-10 & Logistic & No & 0.1 & 0.36  \\
  MNIST & 784-500-500-10 & ReLU & Yes & 0.1 & 1.37 \\
  \hline
 \end{tabular}
 \label{tab:1}
\end{table}

\section{Results}
\subsection{PSA of classifier trained on artificial dataset}
We will describe three ways in which the PSA can be superior to the
analysis based on standard sensitivity map.

\begin{figure}[htbp]
 \centering
 \includegraphics[width=10cm]{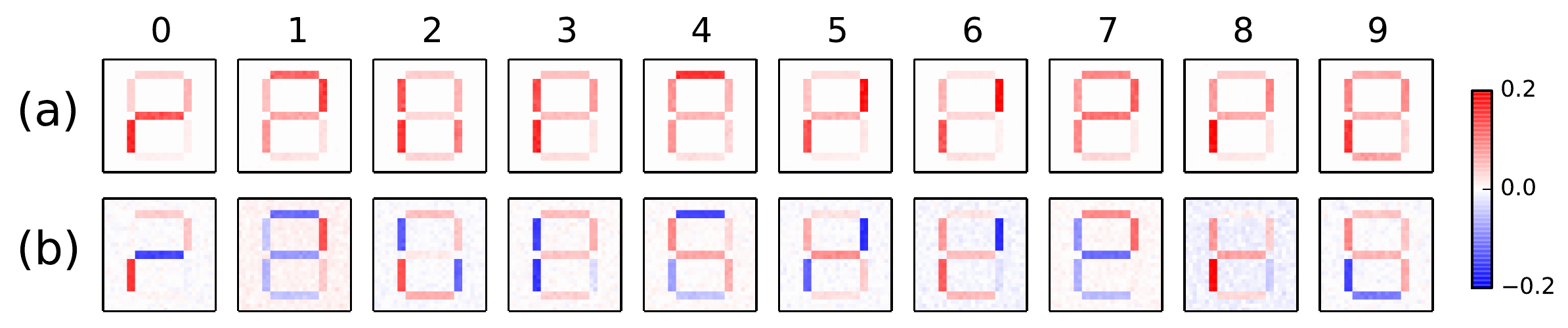}
 \caption{(a) The standard sensitivity maps. (b) The PSMs.
 }
 \label{fig:2}
\end{figure}

Fig.\,\ref{fig:2} compares the PSM and standard
sensitivity map, which were both
obtained from the common neural networks trained for the same
10-class classification problem.
The color intensity of $i$-th pixel represents the magnitude of $v_i$.
Both maps capture the characters that the ``colorless''
rims and likewise ``colorless'' regions enclosed by edges are insignificant in the classification.
Note that the (1st) PSM distinguishes the types of sensitivities by
their sign.
For each numeral, the PSM assigns opposite signs to ``the edges whose \textit{presence} is
crucial in the characterization of the very numeral'' and ``the edges
whose \textit{absence} is crucial in the numeral's characterization.''
This information is not featured in the standard sensitivity map.
For instance, in the sensitivity map for the numeral $1$, the two edges on the
right and the rest of the edges have the opposite sensitivity.
As a result, we can verify the red figure of $1$ in its PSM.
We are able to clearly identify the unbroken figures of $2,4,5$ and $9$ in their
corresponding PSM as well.
We see that, with the extra information
regarding the sign of the sensitivity over each pixel, PSM can provide
us with much richer information than the standard counterpart.

Next, we will show the benefits of sub-principal sensitivity maps
computed from PSA.
Fig.\,3(a) shows the 1st PSM through the 3rd PSM for the numerals $0$ and
$9$.\footnote{We list the PSMs for all the numerals ($0, \dots, 9$) in
the Appendix.}
In order to show how this extra information benefits us in visualization
of the classification problem, we consider the following
``local'' sensitivity map integrated over the samples from a particular
pair of classes:
\begin{align}
\begin{split}
s_{c,c'}(\vec{v})  = \int  \left(\frac{\partial f_c (\vec{x})}{\partial
 \vec{v}} \right)^2 q_{c, c'} (\vec{x}) d\vec{x},
\end{split}  \label{def:l_psm}
\end{align}
where $q_{c, c'}$ is the empirical distribution over the set of samples
generated from the classes $c$ and $c'$.
To get the intuition about this map, note that this value for $(c, c') =
(9, 4)$ can also be pictorially written as
\newsavebox{\cnine}
\newsavebox{\sfour}
\newsavebox{\sfoursmall}
\newsavebox{\pninethird}
\savebox{\cnine}{
  \includegraphics[width=0.45cm]{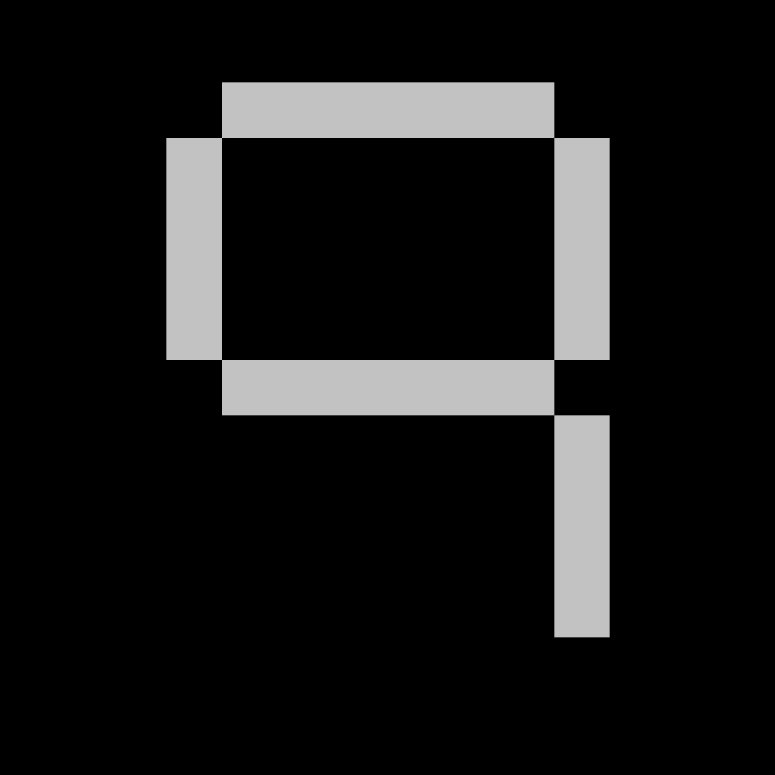}
}
\savebox{\sfour}{
  \includegraphics[width=0.45cm]{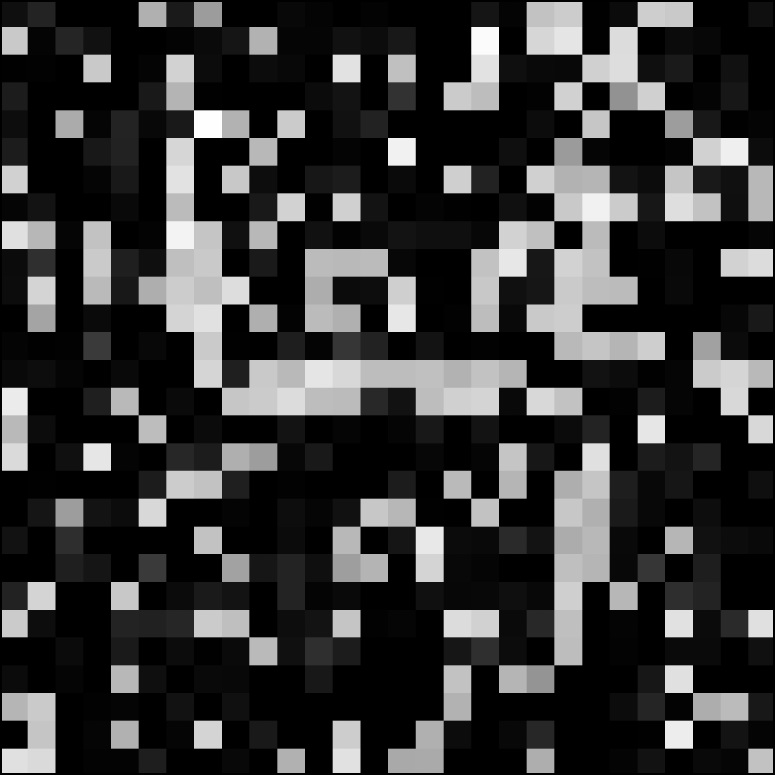}
}
\savebox{\pninethird}{
  \includegraphics[width=0.45cm]{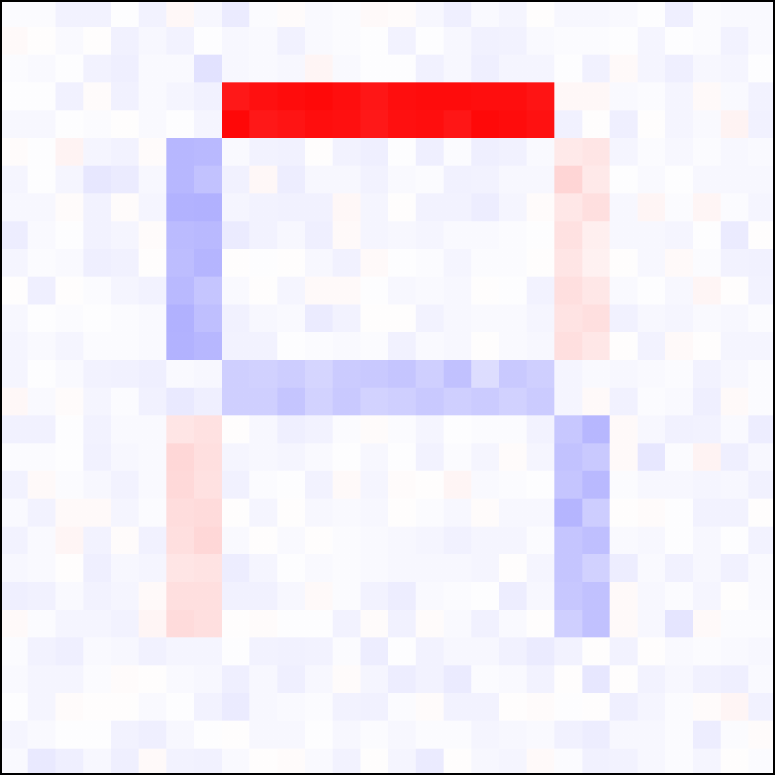}
}
\newlength{\bwcnine}
\settowidth{\bwcnine}{\usebox{\cnine}}
\newlength{\bwsfour}
\settowidth{\bwsfour}{\usebox{\sfour}}
\begin{flalign}
  \lim_{\varepsilon \rightarrow 0}
 E_{\{9, 4\}} \left[
  \left(
  \frac
  {\log P \left(Y = \parbox{\bwcnine}{\usebox{\cnine}} |
  \parbox{\bwcnine}{\usebox{\sfour}} + \varepsilon \vec{v} \right)
  - \log P \left(Y = \parbox{\bwcnine}{\usebox{\cnine}} |
  \parbox{\bwcnine}{\usebox{\sfour}} \right)}
  {\varepsilon}
  \right)^2
  \right],
\end{flalign}
where $\vec{v}$ can be the 3rd PSM of class 9, \usebox{\pninethird}, for
example.
If $\vec{v}_{k}$ is the $k$-th PSM of the classifier,
then $s_{c,c'}(\vec{v}_{k})$ quantifies \textit{the sensitivity of the machine's
answer to the binary classification problem of ``$c$ vs $c'$''} with
respect to the perturbation of the input in the direction of $\vec{v}_{k}$.
By looking at this value for each $k$, we may visualize the ways that the classifier
deals with the binary classification problem.  Such visualization may aid us
in learning from the classifiers the way to distinguish one class from another.
Fig.\,3(b) shows the
values of $s_{c,c'}(\vec{v}_{k})$ for $c \in \{0, 9\}$ and $k \in \{1,\dots,10 \}$.
We could see in the figure that, for the case of
$(c, c') = (9, 4)$, $s_{c,c'}(\vec{v}_{3})$ was larger than
$s_{c,c'}(\vec{v}_{1})$.
This suggests that the 3rd PSM is more helpful than the 1st PSM for
distinguishing $4$ from $9$.
We can actually verify this fact by observing that the 3rd PSM is
especially intense at the top most edge, which can alone differentiate $4$ from $9$.
We are able to confirm many other cases in which the sub-principal sensitivity maps were
more helpful in capturing the characters in binary classification problems than the 1st PSM.
Thus, PSA can provide us with the knowledge of the classifiers
that was inaccessible with the previous method based on the standard
sensitivity map.
\begin{figure}[htbp]
 \begin{minipage}[t]{0.35\columnwidth}
  \centering
   \includegraphics[clip, width=1.\columnwidth]{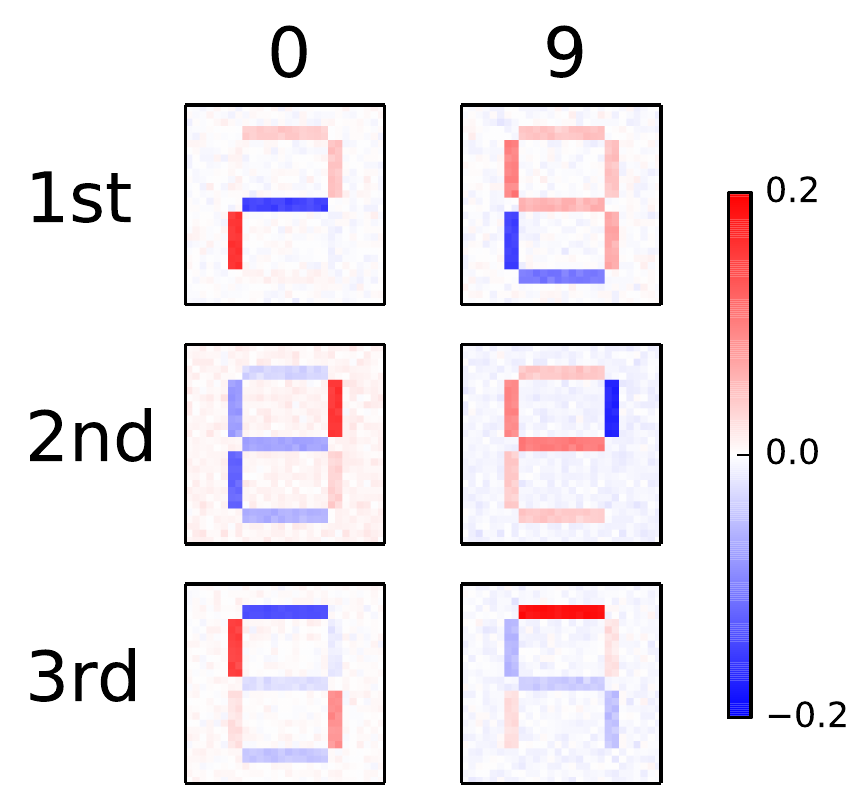}
  \hspace{1.6cm} \textsf{(a)}
  \end{minipage}
 \begin{minipage}[t]{0.6\columnwidth}
  \centering
   \includegraphics[clip, width=1.\columnwidth]{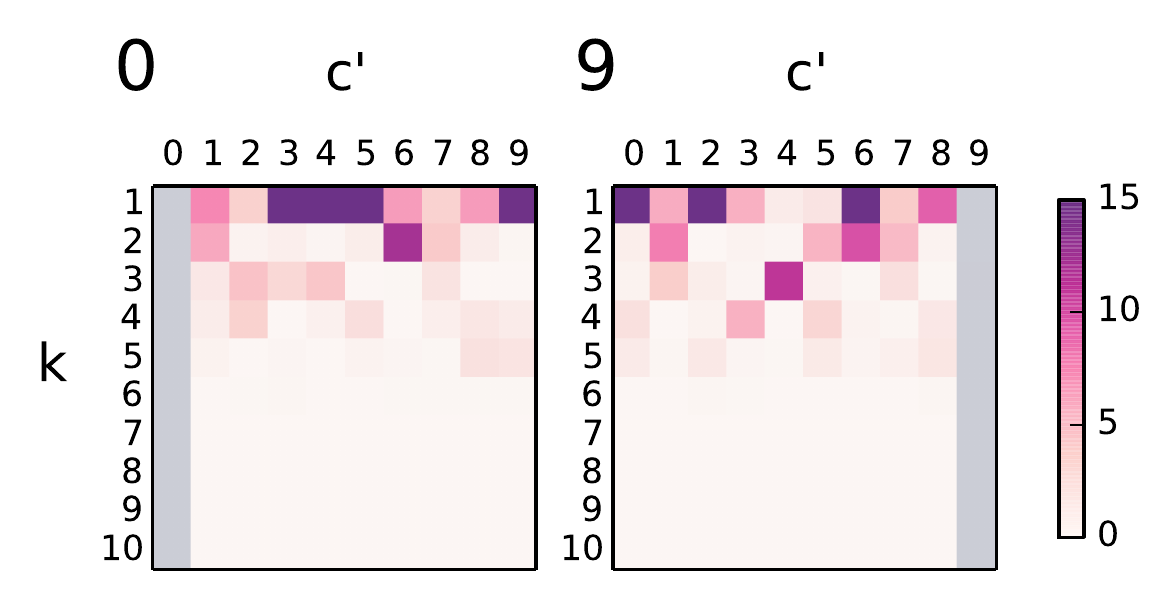}
  \hspace{1.6cm} \textsf{(b)}
 \end{minipage}
  \label{fig:3}
 \caption{(a) 1st $\sim$ 3rd PSMs of the classifier outputs $f_c$ for the numerals
 $0$ and $9$. (b) $s_{c, c'}(\vec{v}_{k})$ for $c \in \{0, 9\}$, $k \in
 \{1, \dots, 10\}$, and $c' \in \{0, \dots, 9\} \backslash \{c\}$.}
\end{figure}

\begin{figure}[htbp]
 \centering
 \includegraphics[width=9cm]{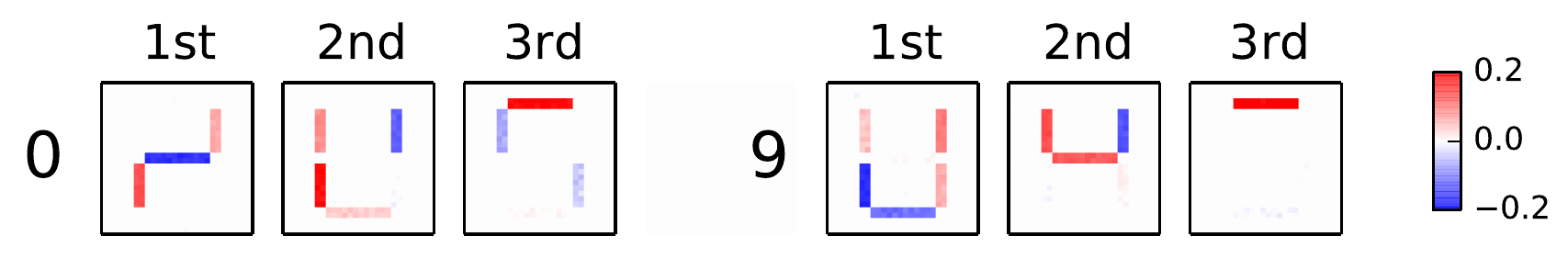}
 \caption{
 Results of the sparse PSA on the classifiers $f_c$ with $p =
 3$ for the numerals $0$ and $9$.
 We ranked the $3$ basis elements by the magnitude of $s(\vec{v})$.
 We selected the regularization term of  $\lambda = 5$,
 and each PSM was normalized so that its $L_2$ norm was $1$.
 }
 \label{fig:4}
\end{figure}

Finally, we demonstrate the usefulness of formulation~\eqref{eq:PSA} in
the construction of sparse and intuitive sensitivity map.
Fig.\,\ref{fig:4} depicts the sensitivity maps obtained from the
application of our sparse PSA in~\eqref{eq:sparsePSA} to the data above.
Note that the sparse PSA not only washes away rather irrelevant
pixels from the canvas, but it also assigns very high intensity to essential pixels.
With these ``localized'' maps, we can better understand the discriminative features
utilized by the trained classifiers.
\subsection{PSA of classifier trained on MNIST dataset}
\begin{figure}[H]
 \centering
 \includegraphics[width=9cm]{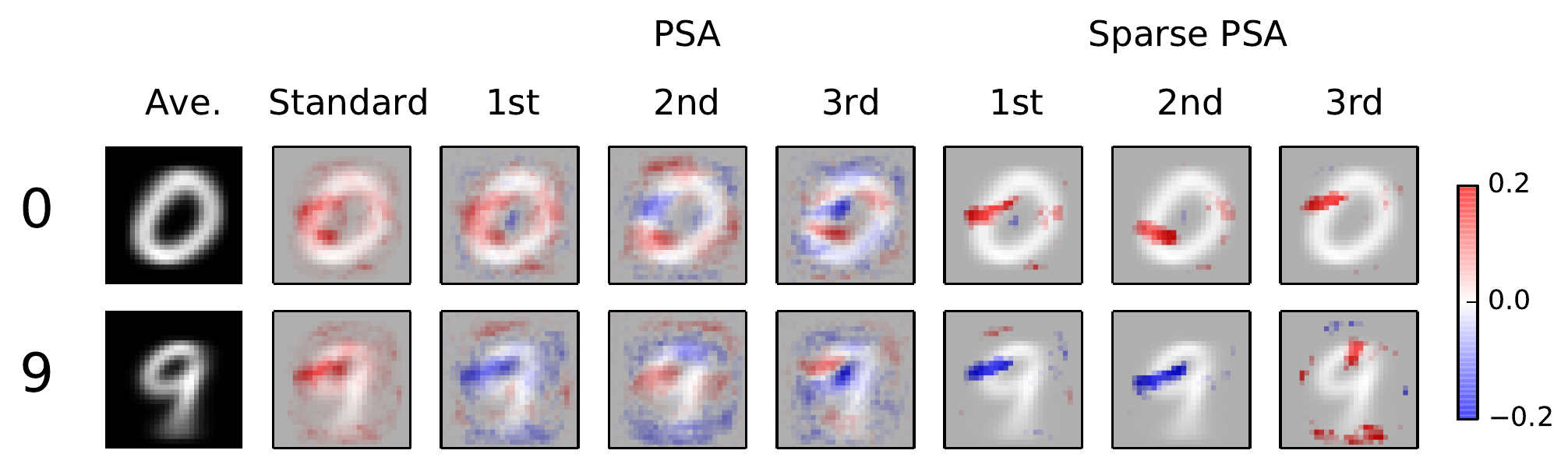}
 \caption{Standard sensitivity map, PSA, and sparse PSA for
 $c \in \{0, 9\}$, $k \in \{1, 2, 3\}$, and $c' \in \{0, \dots
 ,9\} \backslash \{c\}$.
 Ave. stands for the average of the testing dataset for the corresponding numerals.}
 \label{fig:5a}
\end{figure}

We trained a nonlinear neural network-based classifier on the MNIST
dataset, which consists of hand-written digits from $0$ through $9$.
We then analyzed the trained classifier with our PSA.
This dataset illuminates a particular challenge to be confronted in the application of the PSA.
By default, hand-written objects do not share common displacement and orientation.
Without an appropriate registration of input space,
the meaning of each pixel can vary across the samples, making the
visualization unintuitive.
%
%
This is typical in some of the real-world classification problems.
%
In the fields of applied science, standard registration procedure is
often applied to the dataset before the construction of the classifiers.
For example, in neuroimaing, one partitions the image data into anatomical
regions after registration based on the standard brain,
and represents each one of them by a group of voxels.
In other areas of science, one does not necessarily have to face such problems.
In genetics, data can be naturally partitioned into genes~\cite{Yukinawa2009}.
Likewise, in meteorology, 3D dataset is often translated into voxel
structures, and a group of voxels may represent geographical region of
specific terrain~\cite{Kontos2005}.
In this light, the digit recognition in unregistered MNIST data may not
be an appropriate example for showing the effectiveness of our visualization method.
For the reason that we will explain later, registration of multiclass
dataset like MNIST can be difficult.
We chose MNIST dataset here because it is familiar in the community of machine learning.
Fig.\,\ref{fig:5a} summarizes the results.
Both the standard sensitivity map and the PSM were able to
capture the character that outer rims are rather useless in the classification.

\begin{figure}[htb]
 \centering
 \includegraphics[width=7cm]{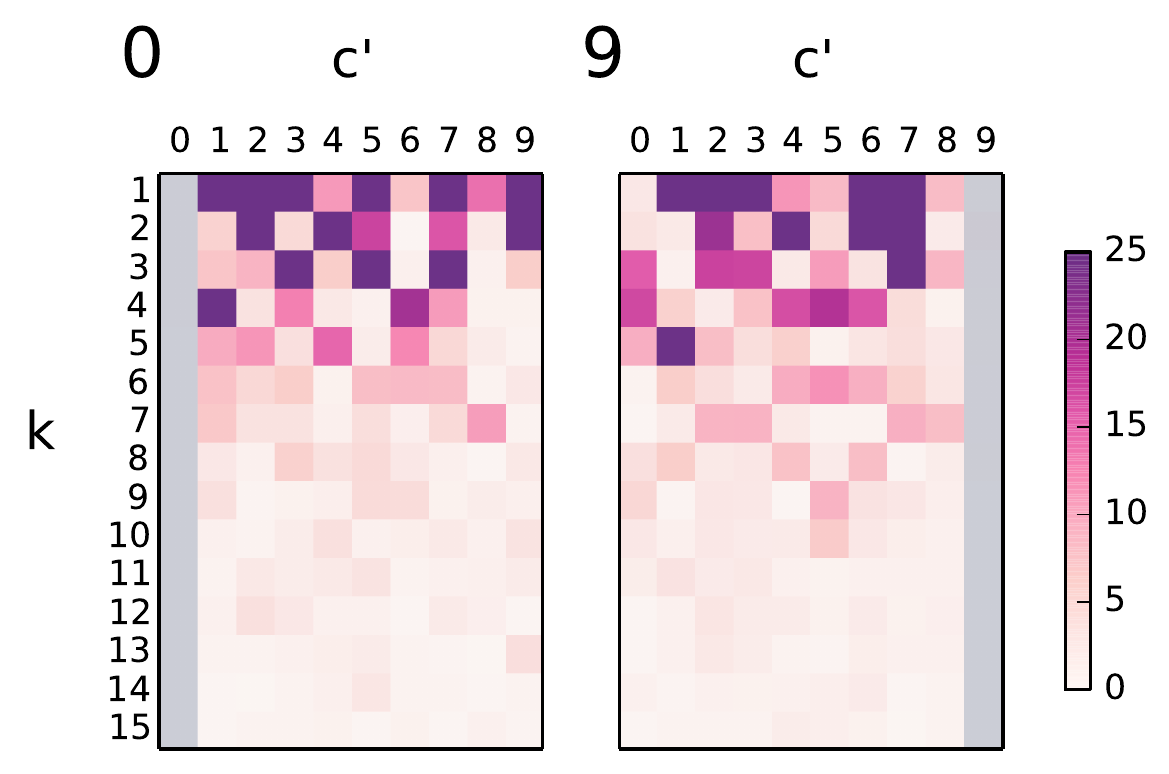}
 \caption{$s_{c, c'}(\vec{v}_{k})$ for $c \in \{0, 9\}$, $k \in
 \{1, \dots, 15\}$, and $c' \in \{0, \dots, 9\} \backslash \{c\}$.}
 \label{fig:5b}
\end{figure}

%
Fig.\,\ref{fig:5b} shows the values of $s_{c,c'}(\vec{v}_{k})$.
We can verify that small numbers of PSMs are complementing each other
in their contributions to the binary classifications.

We also applied sparse PSA to the classifier with $p = 3$ and $\lambda = 40$ (Fig.\,\ref{fig:5a}).
We see that the sparse PSA highlights the essential pixels much more clearly
than the normal PSA.

Since the orientation and position of each numeral pattern
varies across the samples in this dataset,
input dimensions hold different meanings in different samples.
To perform more effective visualization,
we would need registration to adjust each numeral pattern
to a common standard template.
This problem might not be straightforward, since one must prepare
different templates for different numeral patterns.
An elegant standardization suitable for our PSA-based visualization
remains as a future study.
%
%

\section{Discussion}
We proposed a method to decompose the input space based
on the sensitivity of classifiers.
We assessed its performance on classifiers trained with
artificial data and MNIST data.
The visualization achieved with our PSA reveals
at least two general aspects of the classifiers trained in this experiment.
First, note in Fig.~3(b) and Fig.\,\ref{fig:5b} that the first few ($\sim 10$) 
PSMs of the trained classifier dominate the sensitivity
for the binary classification problem.
Second, we see that the classifier use these few PSMs out of $784$
dimensions to solve different binary classification problems.
We are thus able to see that the nonlinear classifiers of the neural
network solve vast number of specific classification problems (such as
binary classification problems) \textit{simultaneously and efficiently}
by tuning its sensitivity to the input in a data-driven manner.
One cannot attain this information with the standard sensitivity map~\cite{Zurada1994,Zurada1997,Kjems2002} alone.
With PSA, one can visualize the decomposition of the
knowledge about the input space learnt by the classifier.
From the PSA of efficient classifier, one may obtain a meaningful decomposition of the input
space that can possibly aid us in solving wide variety of problems.
In medical science, for example, PSA might identify a combination of the
biological regions that are helpful in diagnosis.
PSA might also prove beneficial in sciences using voxel based
approaches, such as geology, atmospheric science, and oceanography.

We may incorporate the principle of the PSA into existing standard statistical methods.
A group Lasso analogue of the PSA, which is currently under our development, may enhance the
interpretability of the visualization even further by identifying sets
of voxels with biological organs, geographical location, etc.
By improving its interpretability, PSA and the PSA-like techniques
might significantly increase the applicability of machine learning
techniques to various high-dimensional problems.

\bibliographystyle{splncs}
\bibliography{bib}
\section*{Appendix}
In this section we list the figures that we omitted in the main text.
\begin{figure}[htb]
 \centering
 \includegraphics[width=10cm]{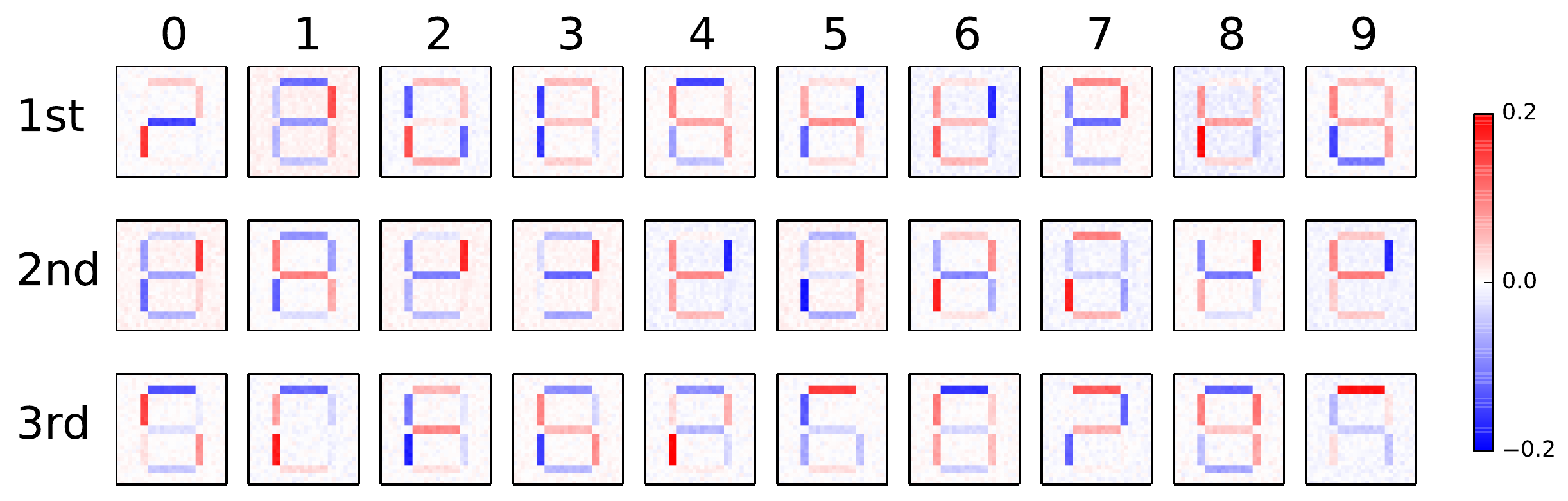}
 \caption{
 1st $\sim$ 3rd PSMs of the classifier trained on the artificial dataset.
 }
 \label{app:1}
\end{figure}
\begin{figure}[htb]
 \centering
 \includegraphics[width=10cm]{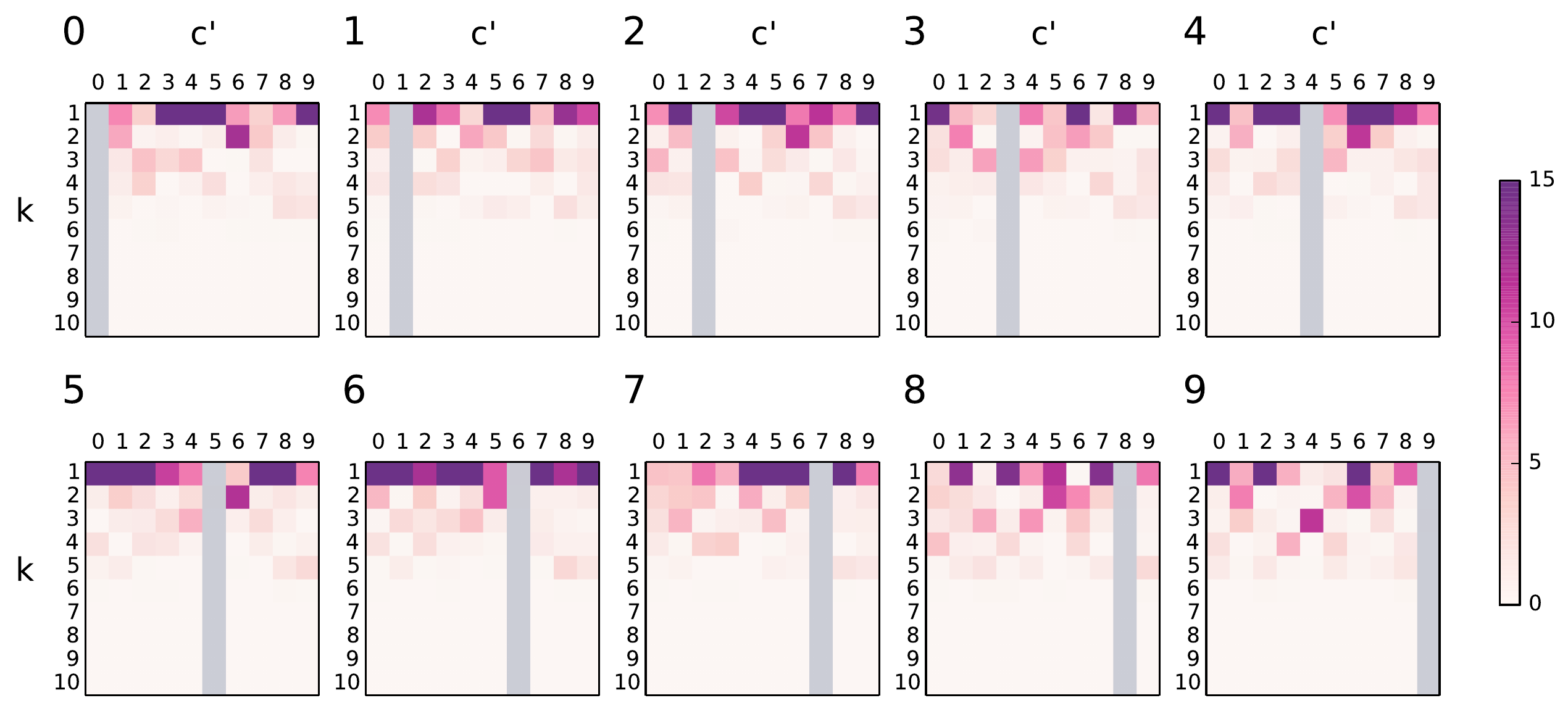}
 \caption{
 $s_{c, c'}(\vec{v}_{k})$ on the artificial dataset.
 }
 \label{app:2}
\end{figure}
\begin{figure}[htb]
 \centering
 \includegraphics[width=10cm]{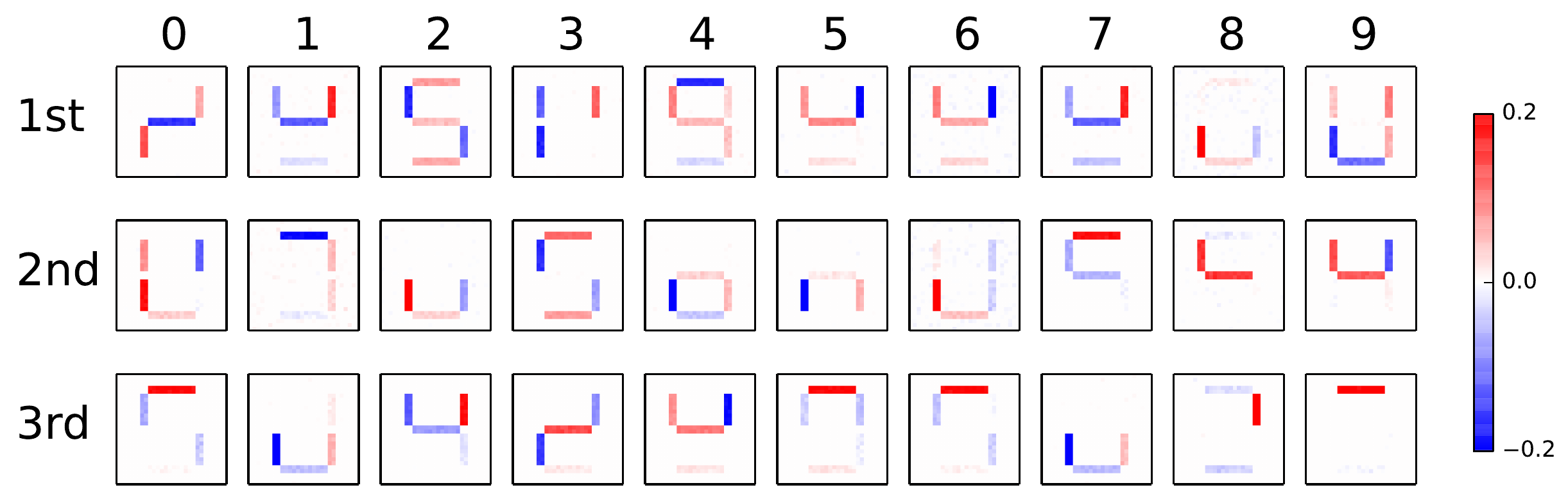}
 \caption{
 Results of the sparse PSA on the classifiers trained on the artificial dataset.
 }
 \label{app:3}
\end{figure}
\begin{figure}[htb]
 \centering
 \includegraphics[width=7.5cm]{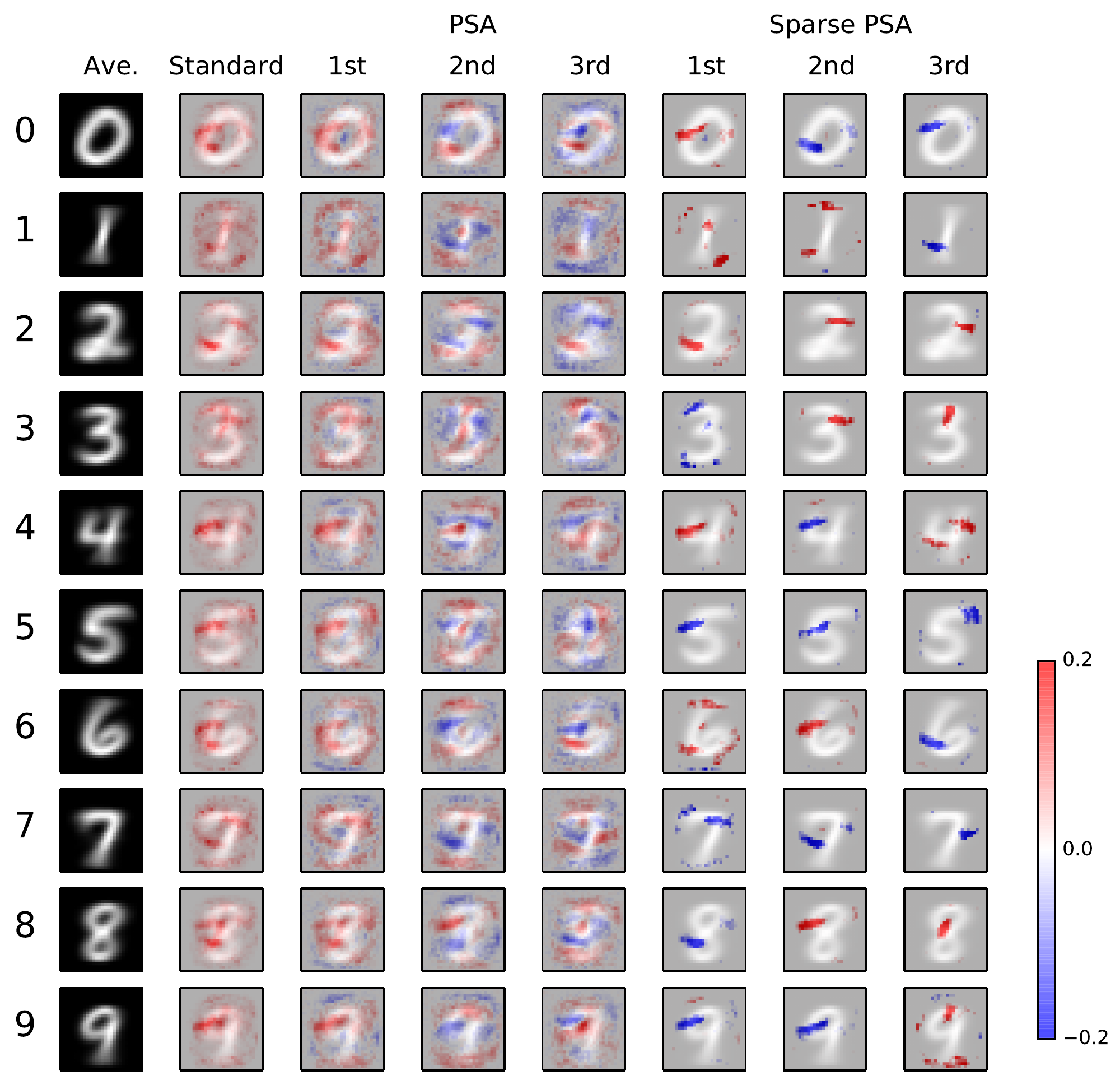}
 \caption{
 Average, standard sensitivity map, PSA, and sparse PSA on MNIST data.
 }
 \label{app:4}
\end{figure}
\begin{figure}[htb]
 \centering
 \includegraphics[width=9.5cm]{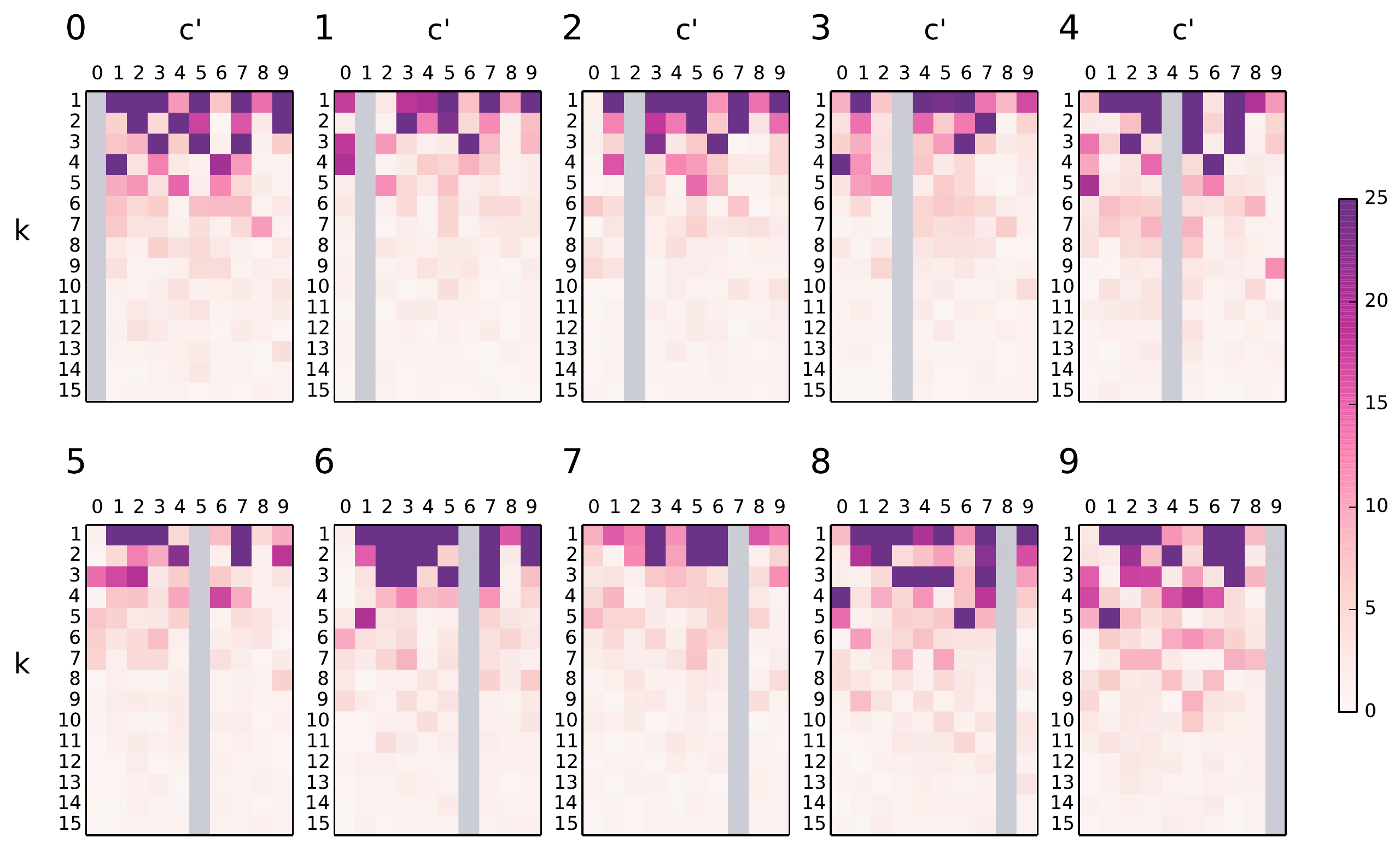}
 \caption{
 $s_{c, c'}(\vec{v}_{k})$ on MNSIT dataset.
 }
 \label{app:5}
\end{figure}

\end{document}